\documentclass{article}

\PassOptionsToPackage{numbers}{natbib}

     \usepackage[preprint]{neurips_2019}

\usepackage[utf8]{inputenc} 
\usepackage[T1]{fontenc}    
\usepackage[hidelinks,colorlinks]{hyperref}
\usepackage{url}            
\usepackage{booktabs}       
\usepackage{amsfonts}       
\usepackage{nicefrac}       
\usepackage{microtype}      

\usepackage[pdftex]{graphicx}
\usepackage{subcaption}
\usepackage{multirow}
\usepackage{threeparttable}

\title{PosNeg-Balanced Anchors with Aligned Features for Single-Shot Object Detection}

\author{%
	Qiankun Tang\textsuperscript{1, 2}\quad Shice Liu\textsuperscript{1, 2}\quad Jie Li\textsuperscript{1, 2}\quad Yu Hu\textsuperscript{1, 2}\\
	\textsuperscript{1}Research Center for Intelligent Computing Systems\\
	Institute of Computing Technology, Chinese Academy of Sciences\\
	\textsuperscript{2} University of Chinese Academy of Sciences\\
	\texttt{\{tangqiankun,liushice,lijie2019,huyu\}@ict.ac.cn} \\ \\
}

\begin{document}
	
	\maketitle
	
\begin{abstract}
We introduce a novel single-shot object detector to ease the imbalance of foreground-background class by suppressing the easy negatives while increasing the positives. To achieve this, we propose an Anchor Promotion Module (APM) which predicts the probability of each anchor as positive and adjusts their initial locations and shapes to promote both the quality and quantity of positive anchors. In addition, we design an efficient Feature Alignment Module (FAM) to extract aligned features for fitting the promoted anchors with the help of both the location and shape transformation information from the APM. We assemble the two proposed modules to the backbone of VGG-16 and ResNet-101 network with an encoder-decoder architecture. Extensive experiments on MS COCO well demonstrate our model performs competitively with alternative methods (40.0\% mAP on \textit{test-dev} set) and runs faster (28.6 \textit{fps}).	
\end{abstract}

\section{Introduction}
\label{sec:intro}
\begin{figure}
	\begin{minipage}[c]{0.5\linewidth}
		\setcaptionwidth{6.5cm}
		\centering
		\includegraphics[width=0.85\textwidth]{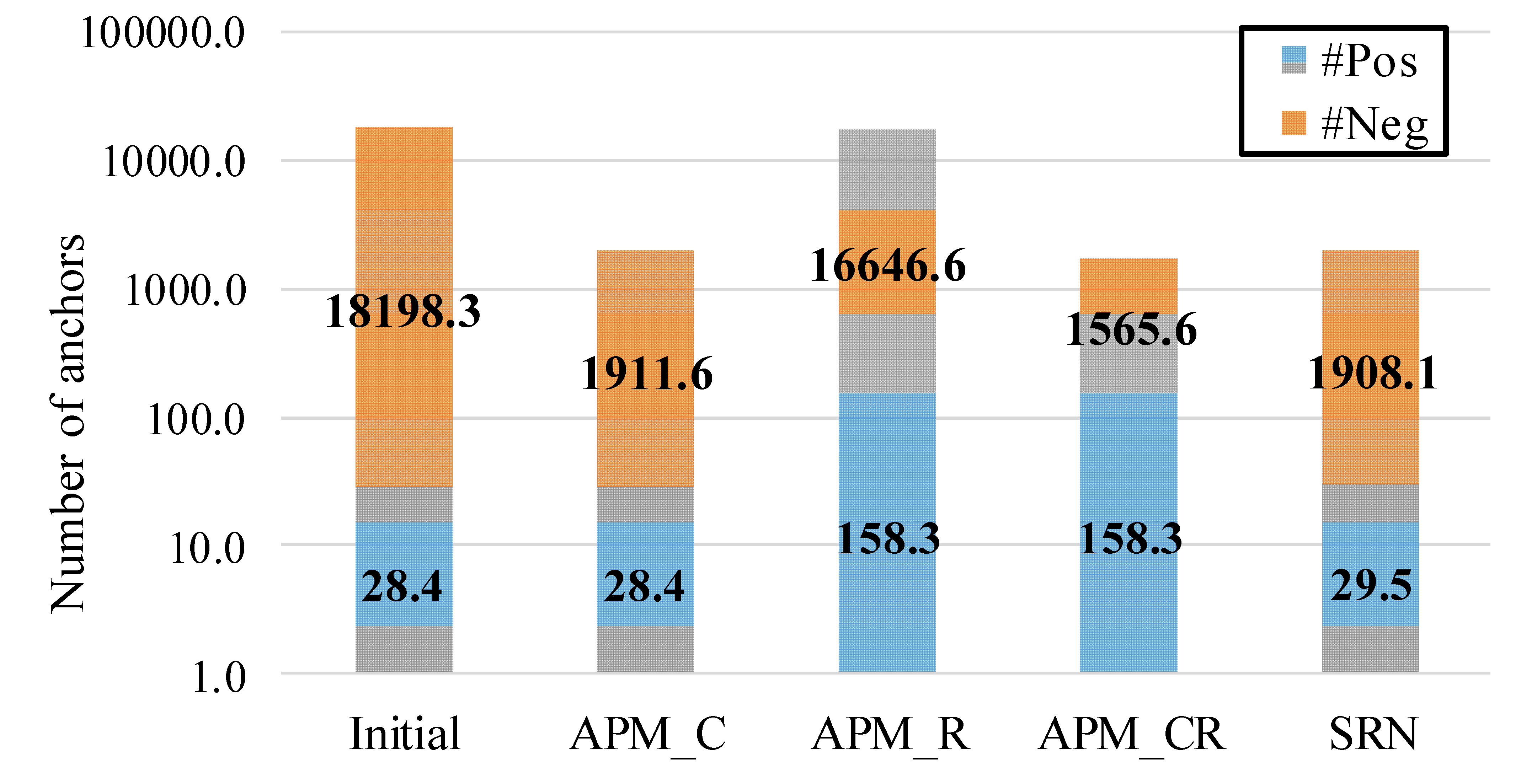}
		\caption{The average number of positive anchors and negative anchors during training at different designs.}
		
		\label{fig:pos_neg_ratio}
	\end{minipage}
	\hfill
	\begin{minipage}[c]{0.5\linewidth}
		\setcaptionwidth{6.5cm}
		\centering
		\includegraphics[width=0.7\textwidth]{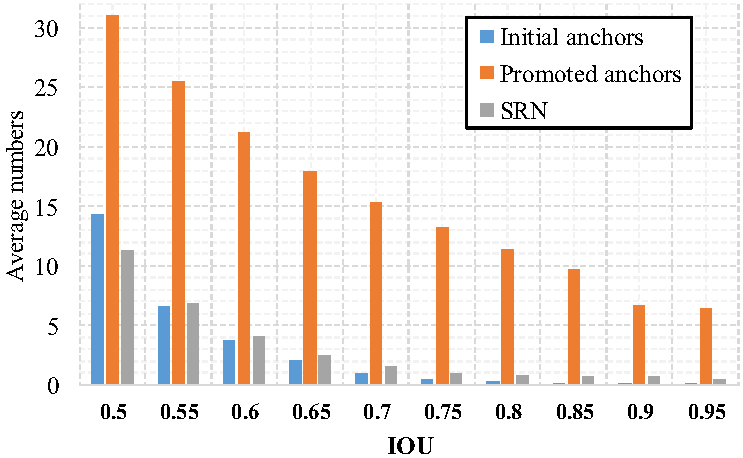}
		\caption{The average number of positive anchors during training at different IoUs before (blue bar) and after (orange bar) promotion.}
		\label{fig:loc_iou}
	\end{minipage}
\vspace{-0.2cm}	 
\end{figure}

The rapid advancements of Convolutional Neural Network (CNN) have largely boosted the development of object detection. Nowadays, the proposal-based approaches~\cite{girshick2014rich,Girshick2015,ren2017faster,lin2017feature,he2017mask} persistently achieve encouraging accuracy. However, they usually suffer from low inference speed. Alternatively, the one-stage framework directly evolves a set of predefined anchor boxes into the final detection results. Although this formulation is simple and enjoys the benefits of high efficiency, its accuracy is lower than that of two-stage detectors. 
One of the main reasons causing this accuracy gap is the imbalance of foreground-background class (shown as the "initial" item in Figure~\ref{fig:pos_neg_ratio}).

To achieve high recall rate, one-stage methods tile plenty of anchors with different scales and aspect ratios over the densely sampled candidate locations of an input image. However, this operation introduces a huge number of background examples. Moreover, the densely tiled anchors usually have poor localization, which results in the reduced number of positives, as shown in Figure~\ref{fig:loc_iou}, the number of initial anchors (blue bar) having intersection-over-union (IoU) with any ground-truth larger than 0.5 decreases rapidly with IoU increasing. Two-stage methods can avoid the class imbalance by generating sparse proposals~\cite{ren2017faster} and  adopting the heuristic sampling to train the detection network. Several solutions have been explored to ameliorate the imbalance issue for one-stage detectors. Kong \etal~\cite{Kong2017} deployed the objectness prior to reduce the searching space of objects. Lin~\etal~\cite{lin2017focal} proposed the Focal Loss, which is a reshaped cross entropy, to down-weight the easy examples and focus training on the hard ones. RefineDet~\cite{zhang2017single} and SRN~\cite{chi2018selective} designed the Anchor Refinement Module and Selective Two-step Classification module separately to identify and filter out the negative anchors before evolving the anchor boxes into detection results. These methods only focus on suppressing easy negatives and ignore the promotion of positive numbers. Although RefineDet and SRN have adjusted the anchors, they just take them to improve the final location accuracy. Meanwhile, in RefineDet, the filtered "negative" anchors may contain the hard positives (about 10\% of the positives) since the scores from Anchor Refinement Module are not confident enough to make a decision. And SRN just adjusted the last few layers which contain only 2.5\% of the anchors. Therefore, their improvements of imbalance are limited. For example, as shown in Figure~\ref{fig:pos_neg_ratio}, SRN decreases much negatives but has marginal improvement of the number of positives as well as their locations (gray bar in Figure~\ref{fig:loc_iou}).

In this work, instead of only suppressing the easy  negatives~\cite{Kong2017,lin2017focal,zhang2017single,chi2018selective}, we solve the imbalance by decreasing the number of easy negatives while increasing the number of positives. Specifically, we propose a module that predicts the probability of each anchor as positive and adjusts their initial locations and shapes. This adjustment can increase the number of positives (the orange bar in Figure~\ref{fig:loc_iou}) by transforming some negative anchors to positives. As such function is analogous to the promotion of spawn to queen in Chess, we name the module as Anchor Promotion Module (APM). During training, the detection module matches all the promoted anchors with the ground-truths and ignores the anchors of negative samples while with low scores of being positive from APM. In this way, the detection module puts more attention to the positive samples and the remaining hard negatives (as shown in the "APM\_CR" item of Figure~\ref{fig:pos_neg_ratio}).

Since the locations and shapes of anchors have been changed as well as their desired information for subsequent detection after the promotion, directly feeding the features associated with initial anchors~\cite{zhang2017single} is sub-optimal. Therefore, we introduce a Feature Alignment Module (FAM) which adaptively encodes necessary features under the guidance of the location and shape transformation information split from the anchor adjustment branch of APM. We adopt the prevalent encoder-decoder architecture with skip-connections~\cite{ronneberger2015u,lin2017feature} as the backbone and attach the proposed modules to each level of the decoder for their rich semantic and detailed information. The overall framework embraces fully convolutional network and is end-to-end trainable.

We demonstrate the effectiveness of our framework, named PADet\footnote{\label{a}Code and models are available at \url{https://github.com/zxhr2793/PADet}{\color{red}}}, by conducting experiments on the challenging benchmark of MS COCO~\cite{lin2014microsoft}. The experimental results show that our framework produces comparable  performance with RestinaNet-101-800~\cite{lin2017focal} but is nearly 7.5$\times$ faster than it. To be specific, with the backbone of VGG-16, the input size of 384$\times$384 achieves 35.2\% mAP on the {\em test-dev} set at the cost of 16 {\em ms} while the size of 512$\times$512 obtains 37.6\% mAP at the cost of 26 {\em ms}.

\section{Related Work}
\label{sec:rw}
{\bf Generic Object Detection}. In general, the CNN-based object detection systems can be roughly divided into two groups. One group is the leading R-CNN-like detectors~\cite{Girshick2015,girshick2014rich,he2015spatial,ren2017faster,dai2016r,lin2017feature,he2017mask,dai17dcn}. They first generate sparsely plausible candidate boxes using the region proposal methods~\cite{ren2017faster,uijlings2013selective}. Then features inside these proposals extracted by RoIPooling~\cite{he2015spatial,Girshick2015} or RoIAlign~\cite{he2017mask} are fed into the second-stage network to refine the coordinates and recognize specific categories. Based on this formulation, the two-stage methods have achieved top performance on several public benchmarks~\cite{everingham2015pascal,lin2014microsoft}. However, they usually take large size images (the shorter side has 800 pixels) as input and complex networks~\cite{he2016deep,xie2016resnext} as backbone which make them very slow to run. The other group is the one-stage detectors which aim at high efficiency by discarding the proposal generation phase and directly predicting categories and locations. YOLO~\cite{redmon2016you} divided the input into several grids and employed a lightweight network to predict confidences and boxes for each cell. YOLO v2~\cite{Redmon2017} further proposed various improvements (BatchNorm~\cite{ioffe2015batch}, dimension clusters, \etc) to YOLO. SSD~\cite{liu2016ssd} and its variants~\cite{liu2018rfbnet,zhang2017single} utilized the pyramidal feature hierarchy to tackle the multi-scale problem. Although they are fast in speed, their performance are largely behind that of two-stage methods. Recently, RetinaNet~\cite{lin2017focal} adopted the ResNet-FPN~\cite{lin2017feature} as backbone and proposed the Focal Loss to address the class imbalance issue. It achieved superior scores even outperformed some two-stage methods but the heavier backbone~\cite{he2016deep} and detection head prevent it from obtaining high efficiency~(only $\sim$5 \textit{fps}~\cite{lin2017focal}). Our work follows the method of one-stage with a main focus on narrowing down the accuracy gap between the two-stage methods and simultaneously maintaining high efficiency.

\noindent{\bf Cascaded Architecture}. Cascaded architecture has been explored a lot for improving classification and refining locations. Viola and Jones~\cite{viola2001rapid} trained a series of cascaded weak classifiers to form a strong region classifier for face detection. MR-CNN~\cite{gidaris2015object} introduced an iterative bounding box regression by feeding the bounding boxes into RCNN several times to improve the localization accuracy during inference. More recently, Cai~\etal~\cite{cai2018cascade} proposed the Cascade R-CNN which achieved more accurate boxes by a sequence of detectors trained with increasing IoU thresholds. Cheng~\etal~\cite{cheng2018revisiting} resampled the hard positive detection boxes and applied a R-CNN to rescore these boxes. Different from the above works which focus on further improving the output detection results in two-stage methods, our framework aims to recognize the positive anchor boxes and promote the anchors for one-stage detection.

\noindent{\bf Feature Alignment}. RoIPooling~\cite{he2015spatial,Girshick2015} is a widely used operation to perform the coarse alignment between proposals and their corresponding features in two-stage methods. RoIAlign~\cite{he2017mask} avoids the spatial quantization of RoIPooling and extracts more aligned representation for boxes and masks. However, they introduce much computational and memory (\eg, at least 1000$\times$ in our settings) burden and thus are not applicable for the one-stage detectors. GA-RPN~\cite{wang2019region} adapts the features for learning proposals by deformable convolution with the offsets learned from the anchor shape prediction branch. While our model captures the aligned features by considering both the location and shape information from the anchor adjustment branch of APM in a more robust manner.

\begin{figure}
	\centering
	\includegraphics[height=4.5cm]{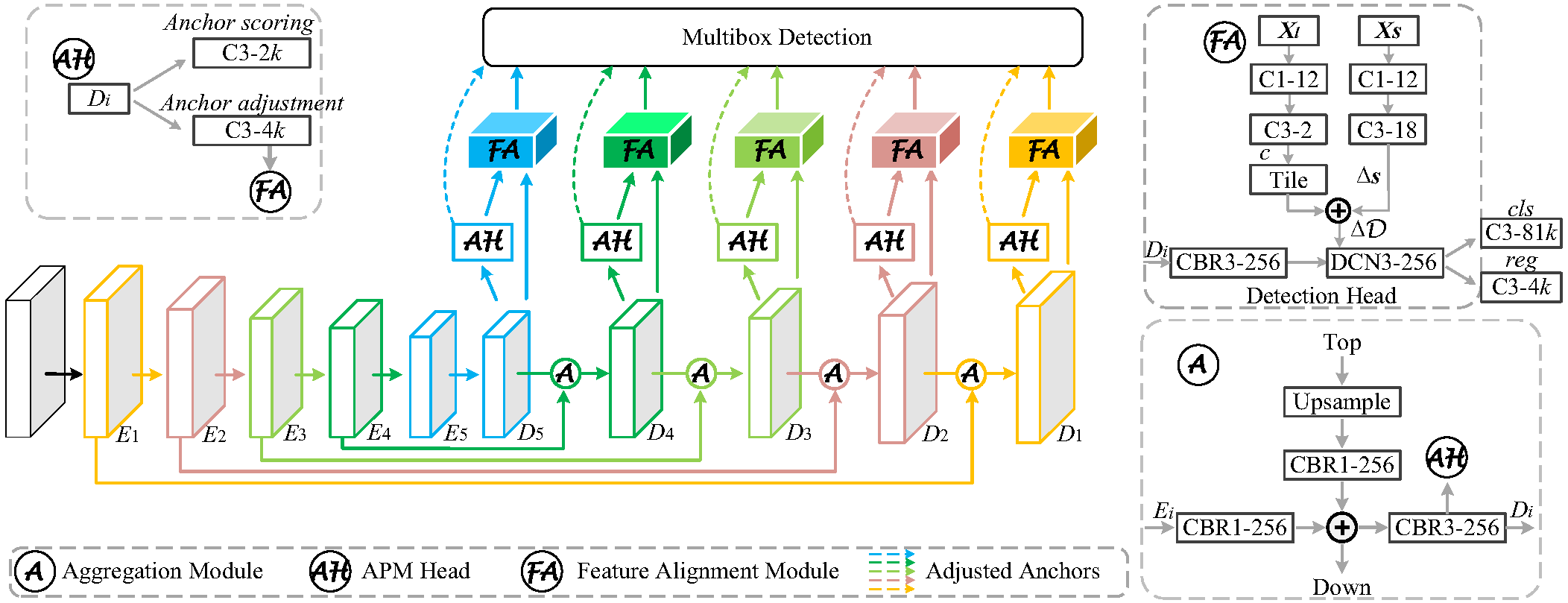}
	\caption{Overview of the proposed framework. The backbone is an encoder-decoder architecture based on VGG-16. On top of each level of the decoder, we attach the proposed anchor promotion module (APM), feature alignment module (FAM) followed by the detection module.}
	\label{fig:framework}
\vspace{-0.2cm}	 	
\end{figure}

\section{Approach}
The encoder-decoder architecture with skip-connections~\cite{ronneberger2015u} has been demonstrated its effectiveness in many computer vision tasks, \eg object detection~\cite{lin2017feature}, semantic segmentation~\cite{peng2017large, zhang2018exfuse}, salient object detection~\cite{Liu2019PoolSal}. Thus, in this work, we adopt this architecture as backbone network. The overall framework is illustrated in Figure~\ref{fig:framework}, we will describe the design of each component in details.

\noindent{\bf Backbone Network}. Similar to SSD~\cite{liu2016ssd}, we take the reduced VGG-16~\cite{simonyan2015vgg}, though other networks are applicable, as base feature extractor. We note the conv4\_3, conv7, conv8\_2, conv9\_2, conv10\_2 blocks as \{$E_1$, $E_2$, $E_3$, $E_4$, $E_5$\} with corresponding strides $s$=\{8, 16, 32, 64, 128\} in regard to the input image. The decoding phase starts from $E_5$ using 1$\times$1 convolutional layer with 256 channels followed by BatchNorm~\cite{ioffe2015batch} and ReLU~\cite{glorot2011deep} layers (denoted as CBR$x$-$y$, $x$ means the kernel size and $y$ is the output channel). Each skip-connection from encoding layers takes the same structure. The decoder progressively upsamples features by bilinear upsampling followed by a CBR1-256 block to normalize features before being fused with the features from encoding layers by element-wise addition, as shown in the bottom-right of Figure~\ref{fig:framework}. We note the decoded layers as \{$D_1$, $D_2$, $D_3$, $D_4$, $D_5$\} corresponding to \{$E_1$, $E_2$, $E_3$, $E_4$, $E_5$\} with the same spatial size. An extra \{$D_6$\} is also added by directly down-sampling the \{$D_5$\} for larger stride. In addition, we insert a CBR3-256 block following each decoded layer to further enhance the representations of fused features. We take two types of size, \ie, 384$\times$384 and 512$\times$512, as input.

\noindent{\bf Anchor Design and Matching}. The anchor boxes in this work follow the settings in SSD. We impose one specific scale anchors with 6 aspect ratios (\eg, 1, 1, 2, 3, 1/2, 1/3) for \{$D_1$, $D_2$, $D_3$, $D_4$\} and 4 aspect ratios of (1, 1, 2, 1/2) for \{$D_5$, $D_6$\} for each feature location. We first match the ground-truths to the anchor boxes with the best IoUs. Then we match the anchor boxes to any ground-truth with IoUs higher than a threshold of 0.5. The anchor box is negative when its IoU with all ground-truths is less than 0.3. We adopt this matching strategy for training both the APM and detection module. 

\begin{figure}
	\centering
	\includegraphics[height=4.8cm]{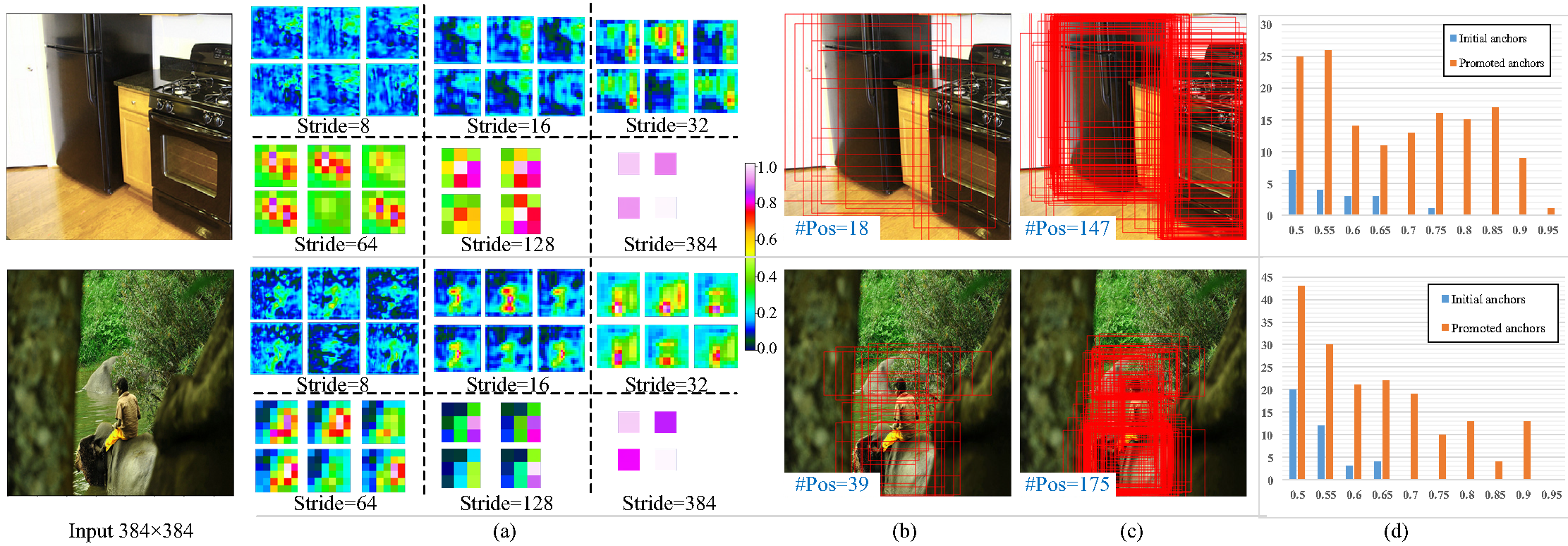}
	\caption{The examples from the outputs of APM. (a) the scores of being positive of anchors at each level, darker color means lower score and lighter is higher. (b) Initial positive anchors, the values on the bottom left are the number of positives. (c) The positive anchors output from APM. (d) The number of positive anchors at different IoU ranges during training before (blue bar) and after (orange bar) being promoted by APM.}
	\label{fig:apm_results}
\vspace{-0.2cm}	 	
\end{figure}

\subsection{Anchor Promotion Module}
Based on the above anchor settings, we calculate the average number of positive anchors and negatives as well as the number of positives at different IoUs with the input image of 384$\times$384. As shown in the "initial" item of Figure~\ref{fig:pos_neg_ratio}, the total anchors are 18400 while positives only take a small fraction or even less. Moreover, the positive anchors have poor initialization of locations (the blue bar in Figure~\ref{fig:loc_iou}). In this way, the negatives will dominate the classifier and regressor leading to suboptimal results. Therefore, it is essential to promote these anchors before forwarding them to the detection module. We propose an Anchor Promotion Module (APM) to achieve this.

 Specifically, as for each anchor box, APM applies a classifier to predict its probability of being positive and a coordinate regressor for coarsely adjusting its initial location and shape. Different from the ARM in RefineDet~\cite{zhang2017single} and detection layers in SSD~\cite{liu2016ssd} being inserted into encoding layers which contain weak semantic information at low-level layers and would misclassify the anchors, we add the proposed APM to the decoding layers, as shown in  Figure~\ref{fig:framework}.

To mitigate the imbalance issue when training the detection module, we apply the match strategy to all the promoted anchor boxes and ignore an anchor if: (i) it is a negative sample and (ii) its score of being positive from APM is lower than a threshold $\theta$ (we empirically set $\theta=0.01$). As a result, parts of the easy negatives will be rejected, we keep the positives and the remaining hard negatives to train the detection module.

As the statistical results shown in Figure~\ref{fig:pos_neg_ratio}, following the above match strategy, if APM only adopts a classifier (APM\_C) for processing the initial anchors, the ratio between positives and negatives is increased by 9.5 times when just rejecting the easy negatives. When APM only being deployed a coordinate regressor (APM\_R), it will output more positives and the ratio is increased from 1:625 to 1:105. The ratio is further increased by 64.7 times to nearly 1:10 when both components are applied (APM\_CR). As shown in Figure~\ref{fig:loc_iou}, compared to the initial settings (blue bar  in Figure~\ref{fig:loc_iou}), both the quality and quantity of positives (orange bar in Figure~\ref{fig:loc_iou}) have been promoted averagely during the training. Figure~\ref{fig:apm_results} presents two examples about the scores of positive anchors across different levels and the number of positives before and after the promotion of APM.

Note that APM is similar to the Region Proposal Network (RPN)~\cite{ren2017faster} but RPN needs the following operations: \textbf{(1)} The outputs from RPN will go through a series of post-processing steps: filtering low-quality proposals, removing duplicate ones. \textbf{(2)} Top $n$ proposals are selected based on their scores for subsequent detection. \textbf{(3)} RoIPooling/RoIAlign is adopted to extract features for the selected proposals. Moreover, there is no interaction between RPN head and detection network after generating proposals. While APM is designed just for promoting the anchors with no need of the above operations and the regression branch of APM can guide the extraction of aligned features as well as the outputs of APM are beneficial to balance the samples for detection module during training.

\subsection{Feature Alignment Module}
In one-stage methods, anchor centers are aligned with feature pixels, so the convolutional features of anchor center are taken as anchor representations. However, after the promotion of APM, the distributions of anchors have been changed significantly. Directly transferring the original features for prediction~\cite{zhang2017single} is sub-optimal. Thereby, we propose a feature alignment module (FAM) to extract aligned features for these promoted anchors.

\begin{figure}
	\begin{minipage}[c]{0.5\linewidth}
		\setcaptionwidth{6.4cm}
		\centering
		\includegraphics[height=1.9cm]{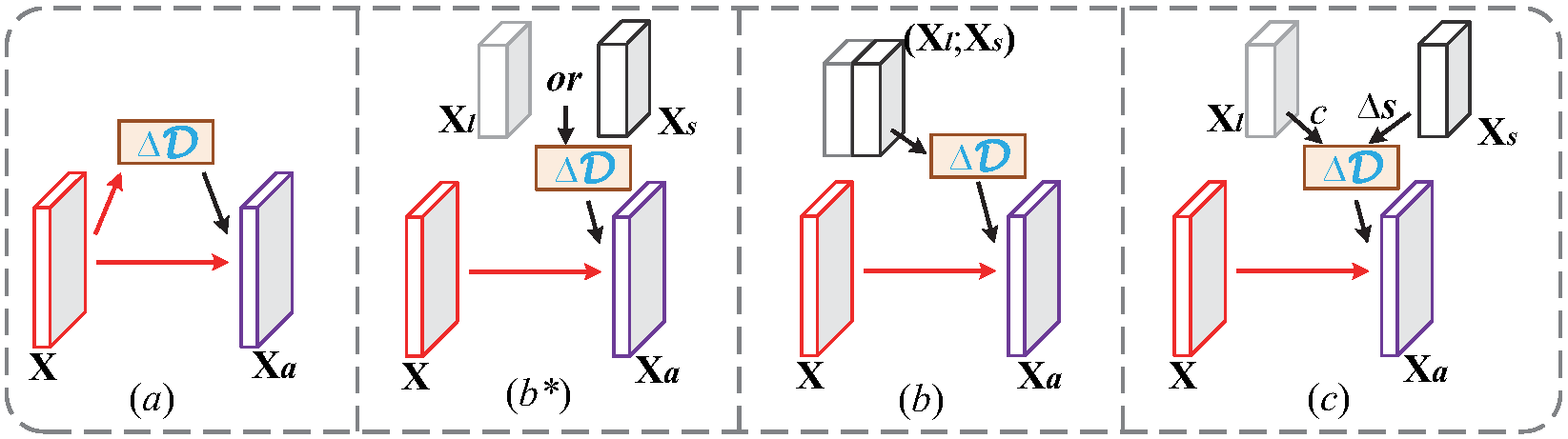}
		\vspace{0.4cm}
		\caption{Different input sources for obtaining offsets.}
		\label{fig:fa}
	\end{minipage} 
	\hfill
	\begin{minipage}[c]{0.5\linewidth}
		\centering		
		\setcaptionwidth{6.5cm}
		\resizebox{\textwidth}{!}{
			\centering
			\begin{tabular}{l|c|ccc|ccc} \toprule
				Option & Variants & AP & $\mbox{AP}_{50}$ & $\mbox{AP}_{75}$ & $\mbox{AP}_{s}$ & $\mbox{AP}_{m}$ & $\mbox{AP}_{l}$   \\ 
				\midrule
				\multicolumn{2}{l|}{w/o alignment } & 31.7 & 53.1 & 33.1 & 16.0 & 35.4 & 45.4 \\
				\midrule
				\multicolumn{2}{l|}{(a)~Implicit } & 32.8 & 54.0 & 34.5 & 16.8 & 36.3 & 46.7  \\
				\midrule
				\multirow{3}*{(b)~Explicit}&$\mathcal{F}$($\mathbf{X}_l$) & 33.6 & 55.0 & 35.7 & 18.2 & 37.4 & 47.4 \\
				~ & $\mathcal{F}$($\mathbf{X}_s$) & 33.4 & 54.6 & 35.0 & 17.6 & 36.8 & 47.8 \\
				~ & $\mathcal{F}$($\mathbf{X}_l$;$\mathbf{X}_s$) & 34.1 & \textbf{55.5} & 36.1 & 17.9 & 37.9 & 47.9 \\
				\midrule
				\multicolumn{2}{l|}{(c)~Disentangled } & \textbf{34.8} & 55.4 & \textbf{37.5} & \textbf{18.6} & \textbf{38.6} & \textbf{49.2} \\
				\bottomrule
			\end{tabular}
		}
	\captionof{table}{Experimental results of different
		choices to acquire offsets.}
	\label{table:fa}	
	\end{minipage}
\vspace{-0.3cm}
\end{figure}

To obtain aligned features, one possible way is to sample the features at the new locations, similar to RoIPooling/RoIAlign, resize and then adopt a filter to extract the anchor representations. We argue that this solution introduces a complex pipeline (sampling$\rightarrow$resizing$\rightarrow$convolution) and it burdens system efficiency and memory. An alternative way, inspired by Deformable ConvNets~\cite{dai17dcn,zhu2018deformable}, is to simultaneously sample and convolve by imposing offsets upon the filter kernels. This is simpler and only involves a few computations and memory (learning offsets and addition). However, one question is how to acquire appropriate offsets (denoted as $\Delta\mathcal{D}$, a vector corresponding to kernels) along with the adjustment of anchors during training? There are several choices as depicted in Figure~\ref{fig:fa}.

{\bf (a) Implicit learning}. A naive way is to learn the offsets implicitly by taking the backbone features as input~(shown as Figure~\ref{fig:fa}~(a)), namely $\Delta\mathcal{D} = \mathcal{F}(\mathbf{X})$, where $\mathbf{X}$ is the backbone features. This is also the standard operation in Deformable ConvNets~\cite{dai17dcn,zhu2018deformable}. However, this would bring limited improvements. The reason behind this is that the offsets learning component is ignorant of the movements of anchors and just implicitly optimized by the detection objectives, while the features are currently misaligned so the objectives only provide suboptimal backward information.

{\bf (b) Explicit learning}. The anchor adjustment branch from APM contains rich location change ($\mathbf{X}_l$) and shape scaling~($\mathbf{X}_s$) information, thus it can provide better guidance for learning offsets. As shown in Figure~\ref{fig:fa}~($\mbox{b}^*$), we can feed the split location change information (\ie, $\Delta\mathcal{D} = \mathcal{F}(\mathbf{X}_l)$) or shape scaling information (\ie, $\Delta\mathcal{D} = \mathcal{F}(\mathbf{X}_s)$) to learn the offsets, but this only brings limited improvement due to the lack of shape or location transformation information. In fact, directly taking the combined $\mathbf{X}_l$ and $\mathbf{X}_s$ (\ie, $\Delta\mathcal{D} = \mathcal{F}(\mathbf{X}_l;\mathbf{X}_s)$ in Figure~\ref{fig:fa}~(b)) as input is also inappropriate. Since $\mathbf{X}_l$ is for translating the locations and $\mathbf{X}_s$ is for altering the shapes of anchors, both their design targets and the scale of contained information are different. The concatenated information may not be utilized sufficiently to result in the optimal offsets $\Delta\mathcal{D}$. 

{\bf (c) Disentangled learning}. Now that the information from the adjustment branch of APM play different roles. We consider factorizing the offsets learning as follows:
$\Delta\mathcal{D}=c + \Delta\mathbf{s}=\mathcal{F}(\mathbf{X}_l) + \mathcal{G}(\mathbf{X}_s)$, where $c$ is a scalar learned from $\mathbf{X}_l$ and is shared by all the filter kernels, $\Delta\mathbf{s}$ is the residual vector learned from $\mathbf{X}_s$, as shown in Figure~\ref{fig:fa}~(c). This formulation is inspired by the intuition that the location change information is helpful for the filters to find the optimal place, it should be easier for the solver to find the perturbations with reference to the place based on the shape scaling information.  

We empirically evaluate the above choices and report their results in {\bf Table}~\ref{table:fa}. The baseline result is obtained without any aligned operations. Among these schemes, option {\bf (c)} is a better choice and achieves the best performance. The top-right of Figure~\ref{fig:framework} displays the detailed architecture of option {\bf (c)}. Its output is of 2$K$ channels for $\Delta \mathbf{s}$ and 2 channels for $c$, where $K$ represents the kernel size ($K$=9 across all our experiments). These two outputs are accumulated for the final offsets and we adopt the DCN~\cite{dai17dcn,zhu2018deformable} to extract the aligned features from the backbone. 

\begin{figure}
	\centering
	\includegraphics[width=14cm]{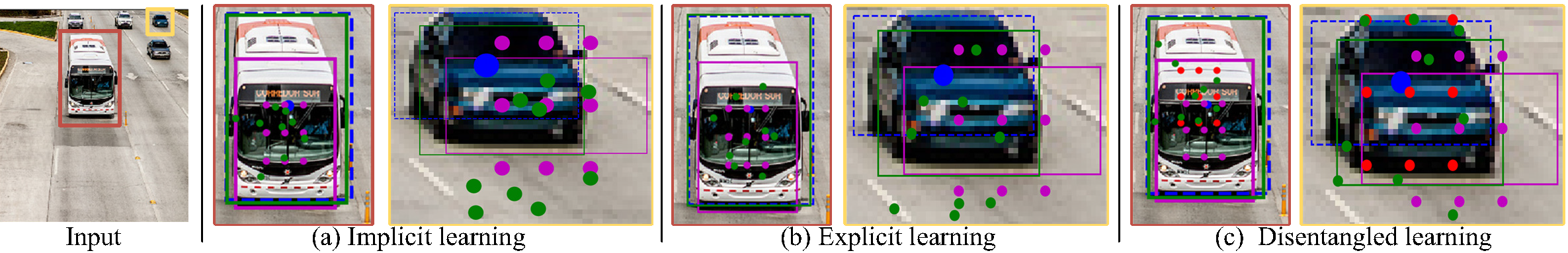}
	\caption{Visualization of the sample locations. Magenta boxes, green boxes and blue boxes represent the initial anchor boxes, promoted anchor boxes and ground truths respectively. The magenta points represent the sample locations of regular convolution. The red points in (c) are the locations of applying $c$ to the kernels. The final sample locations of DCN~\cite{dai17dcn,zhu2018deformable} are in green. Blue points are the centers of ground truth boxes.}
	\label{fig:sample}
	
\end{figure}

We also try to learn the individual offsets for each anchor at a feature location, but the result is just on par with option {\bf (c)}. We conjecture that this is because, during training, parts of the anchors are negatives at a location and the remaining positives usually share the same ground-truth. Thus, the shared offsets at a location is enough to sample necessary features. Moreover, individually learning also costs more computations and memory.

Figure~\ref{fig:sample} displays the sample locations of applying different learning strategies. Adopting the implicit learning will sample around the original locations of regular convolution and extract much irrelevant information especially for the small size objects (\eg, the dark car in Figure~\ref{fig:sample}~(a)). The involved anchor adjustment information indeed helps to sample informative locations in explicit learning. However, the entanglement of location and shape information also leads to the sub-optimal offsets. While the disentangled learning first localizes the appropriate locations (the red points in Figure~\ref{fig:sample} (c)) via $\mathbf{X}_l$, then the information from  $\mathbf{X}_s$ refines these locations to reach optimum.

\subsection{Optimization}
We adopt a lightweight detection head, though a heavier head~\cite{lin2017focal,zhang2017single} can further improve the performance, for high efficiency. As shown in the top-right of Figure~\ref{fig:framework}, the detection head consists of a CBR3-256 block followed by the proposed FAM, then the aligned features are fed into two sibling classification and regression layers. We do not apply the FAM to the last 2 layers for that their feature maps are too small (3$\times$3 and 1$\times$1 for input of 384$\times$384; 2$\times$2 and 1$\times$1 for input of 512$\times$512).
 
\noindent{\bf Loss Function}. The loss function of our model is comprised of two parts: the loss of APM and the loss of detection, which is formulated as follows:
$$L(p, x, c, t) = (1/ N_{apm})(L_b(p, [y\geq1]) + [y\geq1]L_r(x, g)) +(1/N_{d})(L_{cls}(c, y)+[y\geq1]L_r(t, g))$$
where $N_{apm}$ and $N_{d}$ are the numbers of positive sample in APM and detection module. $p$ is the score of being positive. $y$ is the class label and $g$ is the ground-truth box, [$\cdot$] is the indicator function. We use the binary cross-entropy loss $L_b$ for scoring each anchor ($p$) in APM and softmax loss $L_{cls}$ to get specific category score ($c$) in detection. The smooth L1 loss~\cite{Girshick2015} $L_r$ is adopted for adjusting anchor location ($x$) and final detection box ($t$).

\noindent{\bf Training}. The backbone network (VGG-16) are pre-trained on ImageNet. The newly added layers including extra blocks in encoder and decoder layers are initialized by the MSRA~\cite{he2015msra} method. The network is fine-tuned with stochastic gradient descent (SGD) using 0.9 momentum and weight decay of 0.0005. We set the default batch size to 32. During training, we use the data augmentation strategy presented in SSD~\cite{liu2016ssd}, such as horizontal image flipping, image expansion and cropping. Our framework is implemented by PyTorch.

\noindent{\bf Inference}. At inference stage, the APM first outputs the promoted anchors along with their scores of being positive. The anchors are then passed to the detection module for predicting categories and bounding boxes. If the anchors' positive scores lower than the threshold $\theta$, we discard the corresponding refined bounding boxes. Finally, the remaining bounding boxes are post-processed by soft-NMS~\cite{bodla2017soft} ($\sigma$=0.5, $N_t$=0.3) with linear kernel for per class after filtering the confidence scores lower than 0.01 and we take the top 300 detections per image as final results.

\section{Experiments}
We conduct experiments on the challenging benchmark of MS COCO which contains 80 categories. For training, we follow the common practice in~\cite{lin2017feature,lin2017focal} and use the {\em trainval35k} set. All ablative analysis are reported by evaluating on the {\em minival} set. We report the final performance on the {\em test-dev} set. 

\noindent{\bf Implementation Details}. We set the initial learning rate to 2$\times10^{-3}$. In the beginning, outputs from APM are in chaos and make the training unstable. Thus, we apply a "warm up" strategy which gradually ramps up the learning rate from $10^{-6}$ to 2$\times10^{-3}$ for the first 5 epochs. Then we decay the learning rate to 2$\times10^{-4}$ and 2$\times10^{-5}$ at 100 epochs and 140 epochs. The learning ends up at 160 epochs. When the input size is $384\times384$, we train it on 2 GPUs and 4 GPUs for $512\times512$. In the following, we run numerous experiments to analyze the effectiveness of each proposed component at the image size of $384\times384$. We first train the model with backbone of the designed encoder-decoder architecture, this can be regarded as the vanilla DSSD~\cite{fu2017dssd}, and take it as the baseline. Its result is shown in the first row of Table~\ref{tab:effect_overall}.

\subsection{Ablation Study}

\begin{table}
	\centering
	\begin{minipage}[t]{0.45\linewidth}
		\setcaptionwidth{6.5cm}
		\resizebox{\textwidth}{!}{%
			\centering
			\begin{tabular}{ccc|ccc|ccc}
				\toprule
				APM\_C & APM\_R & FA &
				$\mbox{AP}$ & $\mbox{AP}_{50}$ & $\mbox{AP}_{75}$ & $\mbox{AP}_{s}$ & $\mbox{AP}_{m}$ & $\mbox{AP}_{l}$   \\
				\midrule
				~ & ~ & ~ & 29.6 & 49.4 & 30.8 & 12.1 & 32.4 & 44.3\\
				$\checkmark$ & ~ & ~ & 31.0 & 51.3 & 32.6 & 14.3 & 34.5 & 44.9\\			
				~ & $\checkmark$ & ~ & 30.6 & 50.5 & 32.4 & 13.9 & 34.1 & 44.7\\
				$\checkmark$ & $\checkmark$ & ~ & 31.7 & 53.1 & 33.1 & 16.0 & 35.4 & 45.4 \\
				$\checkmark$ & $\checkmark$ & $\checkmark$ & \textbf{34.8} & \textbf{55.4} & \textbf{37.5} & \textbf{18.6} & \textbf{38.6} & \textbf{49.2}\\
				\bottomrule
			\end{tabular}%
		}
	\vspace{0.15cm}
		\caption{The effects of different component designs.}
		\label{tab:effect_overall}
	\end{minipage}
	\hfill
	\begin{minipage}[t]{0.5\linewidth}
		\setcaptionwidth{6cm}
		\centering
		\resizebox{0.8\textwidth}{!}{%
			\centering
			\begin{tabular}{cc|ccccc}
				\toprule
				APM\_R & FA & $\mbox{AP}_{50}$ & $\mbox{AP}_{60}$ & $\mbox{AP}_{70}$ & $\mbox{AP}_{80}$ & $\mbox{AP}_{85}$   \\
				\midrule
				~ & ~ & 49.4 & 44.2 & 36.4 & 24.1 & 16.0\\
				$\checkmark$ & ~ & 53.1 & 47.9 & 39.1 & 25.3 & 16.5\\
				$\checkmark$ & $\checkmark$ & \textbf{54.7} & \textbf{49.7} & \textbf{41.8} & \textbf{28.6} & \textbf{19.2}\\
				\bottomrule
			\end{tabular}%
		}
	\vspace{0.2cm}
		\caption{The effects of anchor adjustment across different IoUs.}
		\label{tab:effect_iou}
	\end{minipage}
\vfill
\begin{minipage}[c]{0.45\linewidth}
	\setcaptionwidth{6cm}
	\resizebox{\textwidth}{!}{%
		\centering
		\begin{tabular}{c|c|cccccc}
			\toprule
			Method & Area &
			$\mbox{AP}$ &
			$\mbox{AP}_{50}$ & $\mbox{AP}_{60}$ & $\mbox{AP}_{70}$ & $\mbox{AP}_{80}$ & $\mbox{AP}_{85}$   \\
			\midrule
			~ & small & 12.1 & 24.6 & 20.1 & 14.2 & 6.9 & 3.8\\
			Baseline & medium & 32.4 & 56.3 & 50.0 & 40.5 & 24.4 & 14.7\\
			~ & large & 44.3 & 67.7 & 62.4 & 54.1 & 41.3 & 30.0\\
			\midrule
			~ & small & 18.6 & 35.1 & 29.8 & 22.9 & 12.4 & 6.6\\
			Ours & medium & 38.6 & 62.6 & 57.1 & 48.9 & 32.5 & 19.9\\
			~ & large & 49.2 & 70.7 & 66.8 & 60.7 & 48.1 & 35.8\\
			\bottomrule
		\end{tabular}%
	}
\vspace{0.15cm}
	\caption{Location accuracy for various areas across different IoUs.}
	\label{tab:effect_eara_iou}
\end{minipage}
\hfill
\begin{minipage}[c]{0.45\linewidth}
	\setcaptionwidth{6.5cm}
	\centering
	\resizebox{0.65\textwidth}{!}{%
		\centering
		\begin{tabular}{c|c|c|c}
			\toprule
			
			Method & Backbone & Input size & mAP \\
			\midrule
			Faster~\cite{he2016deep} & ResNet-101 & $\sim$1000$\times$600 & 76.4\\			
			R-FCN~\cite{dai2016r} & ResNet-101 & $\sim$1000$\times$600 & 80.5\\
			DCR~\cite{cheng2018revisiting} & ResNet-101 & $\sim$1000$\times$600 & 82.5 \\
			\midrule
			SSD~\cite{liu2016ssd} & VGG-16 & 512$\times$512 & 79.8\\
			RefineDet~\cite{zhang2017single} & VGG-16 & 512$\times$512 & 81.8\\
			RFBNet~\cite{liu2018rfbnet} & VGG-16 & 512$\times$512 & 82.2 \\
			PFPnet-R~\cite{kim2018pfpnet} & VGG-16 & 512$\times$512 & 82.3\\ 
			\midrule
			Ours & VGG-16 & 384$\times$384 & 82.0\\
			Ours & VGG-16 & 512$\times$512 & \textbf{82.7}\\
			\bottomrule
		\end{tabular}%
	}
\vspace{0.15cm}
	\caption{Detection results on PASCAL VOC.}
	\label{tab:effect_voc}
\end{minipage}
\vspace{-0.5cm}
\end{table}

\noindent{\bf Effectiveness of APM}. We first add the APM with only anchor scoring branch~(APM\_C) to the baseline. As shown in Table~\ref{tab:effect_overall}, only filtering out the easy negative anchors, the mAP is improved from 29.6\% to 31.0\%. Although the anchor scoring branch eases the imbalance issue to some extent, the poor initialization of anchors impacts the results. Next, we evaluate the impact of APM with only anchor adjustment branch~(APM\_R). The third row of Table~\ref{tab:effect_overall} shows that the APM\_R improves on the baseline, but the imbalance situation results in its limited improvements. When the full APM is adopted, the mAP is able to be further improved. However, this only brings 2.1\% higher mAP due to the lack of aligned features for detection.

\noindent{\bf Feature Alignment}. Adding the feature alignment module improves the performance by 5.2\% mAP, which indicates the importance of alignment between the features and anchors. The aligned features contain fitting information for training the promoted anchors. As analyzed above, the thoughtful design of obtaining offsets for deformable convolution contributes a lot to this improvement.

\noindent{\bf Localization Accuracy}. We also compute the mAP at different IoU thresholds to investigate the impact of anchor promotion on localization accuracy. Table~\ref{tab:effect_iou} shows that anchor promotion consistently boosts the detection accuracy, especially when equipped with the aligned features. However, we also observe that the improvements on higher IoUs (>0.7) are smaller than that of lower IoUs, \eg, 5.4\% of $\mbox{AP}_{70}$ vs. 3.2\% of $\mbox{AP}_{85}$. We guess this is due to the loose hypotheses of positives (IoU $\ge$ 0.5) we adopted for training detection module, the detector is suboptimal to the higher quality results~\cite{cai2018cascade}. Cascading multiple detection modules with feature alignment modules using improved hypotheses, as in~\cite{cai2018cascade}, may be helpful, we will explore the solutions for this issue in the future work.  

\noindent{\bf Detection Accuracy of Different Scales}. In Table~\ref{tab:effect_eara_iou}, we also investigate the detection accuracy for different areas. As for objects of medium and large areas, their improvements are progressively increased across the IoU ranges of 0.5 to 0.8. We improve the small sizes by a large margin from 12.1\% of mAP to 18.6\%. Nevertheless, the improvements of small objects decrease rapidly with IoU increasing in that only 34.3\% ground-truth small objects (areas < $32^2$) have more than 2$\times$2 pixels on the highest resolution (stride=8) of feature map which we can take for detection. 

\noindent{\bf Results on PASCAL VOC}. To further verify the effectiveness of our model, we conduct experiments on the PASCAL VOC 2007~\cite{everingham2015pascal} dataset. We report the results in Table~\ref{tab:effect_voc}. Our model achieves competitive or even better results compared to some two-stage methods with ResNet-101 and outperforms the recent state-of-the-art one-stage methods.

\subsection{Comparison to State of the Art}
We compare our results on the MS COCO {\em test-dev} set with other state-of-the-art results in Table~\ref{tab:test_compare}. It is noteworthy that, with only VGG-16 and input of $384\times384$, we achieve mAP of 35.2\% without multi-scale training and any testing tricks, which is on par with the RetinaNet700 with ResNet50-FPN as backbone. The accuracy of our method can be further improved to 37.6\% by using a larger input of $512\times512$, which is a competitive result with RetinaNet-101-800. In the fourth column of Table~\ref{tab:test_compare}, we also present the inference speed of different models. Our speed is measured on a machine with Nvidia GTX 1080 Ti, PyTorch 0.4.1, i7-6850k CPU, CUDA 9.0 and cuDNN v7. It is clear that our method performs more efficiently with promising accuracy based on our compact but effective model.

\begin{table}
	\centering
	
\resizebox{0.8\textwidth}{!}{%
	\centering
	\begin{tabular}{l|c|c|c|ccc|ccc}
					\toprule
					~ & Backbone & Input size & FPS &
					$\mbox{AP}$ & $\mbox{AP}_{50}$ & $\mbox{AP}_{75}$ & $\mbox{AP}_{s}$ & $\mbox{AP}_{m}$ & $\mbox{AP}_{l}$ \\
					
					\midrule
					\textit{Multi-shot methods} & ~ & ~ & ~ & ~ & ~ & ~ & ~ & ~ & ~\\
					Faster~\cite{ren2017faster} & VGG-16 & $\sim$1000$\times$600 & 7.0 & 21.9 & 42.7 & - & - & - & -\\
					R-FCN~\cite{dai2016r} & ResNet-101 & $\sim$1000$\times$600 & 9.0 & 29.9 & 51.9 & - & 10.8 & 32.8 & 45.0\\			
					FPN~\cite{lin2017feature} & ResNet-101 & $\sim$1333$\times$800 & 6.0 & 36.2 & 59.1 & 39.0 & 18.2 & 39.0 & 48.2\\
					Mask-RCNN~\cite{he2017mask} & ResNet-101-FPN & $\sim$1333$\times$800 & 5.1 & 38.2 & 60.3 & 41.7 & 20.1 & 41.1 & 50.2 \\
					Cascade-RCNN~\cite{cai2018cascade} & ResNet-101-FPN & $\sim$1333$\times$800 & 7.1 & 42.8 & 62.1 & 46.3 & 23.7 & 45.5 & 55.2 \\
					DCNv2~\cite{zhu2018deformable} & ResNet-101-FPN & $\sim$1333$\times$800 & - & 44.0 & 65.9 & 48.1 & 23.2 & 47.7 & 59.6 \\
					\midrule
					\textit{Single-shot methods} & ~ & ~ & ~ & ~ & ~ & ~ & ~ & ~ & ~\\
					SSD~\cite{liu2016ssd,fu2017dssd} & ResNet-101 & 513$\times$513 & - & 31.2 & 50.4 & 33.3 & 10.2 & 34.5 & 49.8 \\
					DSSD~\cite{fu2017dssd} & ResNet-101 & 513$\times$513 & 5.5 & 33.2 & 53.3 & 35.2 & 13.0 & 35.4 & 51.1 \\
					RetinaNet700~\cite{lin2017focal} & ResNet-50-FPN & $\sim$700$\times$700 & 8.3 & 35.1 & 54.2 & 37.7 & 18.0 & 39.3 & 46.4 \\
					RetinaNet800~\cite{lin2017focal} & ResNet-101-FPN & $\sim$1280$\times$800 & 5.1 & 37.8 & 57.5 & 40.8 & 20.2 & 41.1 & 49.2 \\
					RefineDet~\cite{zhang2017single} & VGG-16 & 512$\times$512 & 24.1 & 33.0 & 54.5 & 35.5 & 16.3 & 36.3 & 44.3 \\
					RFBNet~\cite{liu2018rfbnet} & VGG-16 & 512$\times$512 & 30.3 & 34.4 & 55.7 & 36.4 & 17.6 & 37.0 & 47.6 \\
					PFPNet-R~\cite{kim2018pfpnet} & VGG-16 & 512$\times$512 & 22.2 & 35.2 & 57.6 & 37.9 & 18.7 & 38.6 & 45.9 \\
					CornerNet~\cite{law2018cornernet} & Hourglass-104 & 511$\times$511 & 4.1 & 40.5 & 56.5 & 43.1 & 19.4 & 42.7 & 53.9 \\
					GA-RetinaNet~\cite{wang2019region} & ResNet-50-FPN & $\sim$1280$\times$800 & - & 37.1 & 56.9 & 40.0 & 20.1 & 40.1 & 48.0 \\
					FSAF~\cite{zhu2019fsaf}(\textit{multi-scale}) & ResNet-101-FPN & $\sim$1280$\times$800 & 5.6 & 40.9 & 61.5 & 44.0 & 24.0 & 44.2 & 51.3 \\
					\midrule
					Ours & VGG-16 & 384$\times$384 & \textbf{62.5} & 35.2 & 55.9 & 38.1 & 17.7 & 38.2 & 48.3\\
					Ours & VGG-16 & 512$\times$512 & \textbf{38.5} & 37.6 & 58.7 & 41.0 & 21.0 & 40.4 & 49.5 \\
					Ours & ResNet-101 & 384$\times$384 & 40.0 & 36.9 & 57.0 & 40.2 & 16.4 & 40.4 & 53.3 \\
					Ours & ResNet-101 & 512$\times$512 & 28.6 & 40.0 & 60.8 & 43.8 & 21.3 & 43.9 & 53.9 \\
					\bottomrule
				\end{tabular}%
			}
		\vspace{0.15cm}
\caption{Detection comparisons on MS COCO {\em test-dev} set. \textit{multi-scale} means multi-scale training.}
\label{tab:test_compare}
\vspace{-0.55cm}	
\end{table}
\section{Conclusion}
In this work, we ameliorate the imbalance of  foreground-background class during training the one-stage methods by decreasing the number of easy negatives while increasing the positives. We propose an Anchor Promotion Module (APM) to jointly predict the probability of being positive and adjust the location and shape for each anchor. In this way, we are able to promote both the quality and quantity of positive anchors for training the detection module. We further design an efficient feature alignment module (FAM) to extract the fitting features for these promoted anchors based on the location and shape transformation information from the adjustment branch of APM. These advanced components enable us to achieve attractive performance while keeping high-efficiency.

\bibliographystyle{unsrtnat}
\bibliography{egbib}

\end{document}